\documentclass[12pt]{article}
\usepackage{amsmath}
\usepackage{graphicx,psfrag,epsf}
\usepackage{enumerate}
\usepackage{natbib}
\usepackage{url} % not crucial - just used below for the URL 

\usepackage{xcolor}
\usepackage{amsfonts}
\usepackage{amssymb}
\usepackage{amsthm}

\usepackage{bm}
\usepackage[utf8]{inputenc}
\usepackage{soul}

\usepackage{ulem}

\usepackage{pdflscape}
\usepackage{bbm}
\usepackage{subfig}
\usepackage{algorithm}
\usepackage{algorithmic}

\usepackage[colorlinks=true,
            linkcolor=blue,
            urlcolor=blue,
            citecolor=blue]{hyperref}

\usepackage[english]{babel}
\newcommand\mat[1]{\begin{pmatrix}#1\end{pmatrix}}
\usepackage{bigints}

\usepackage{float}
\usepackage{subfig}

\newtheorem{Theorem}{Theorem}
\newtheorem{Corollary}{Corollary}
\newtheorem{Lemma}{Lemma}

\newtheorem{condition}{Condition}

\newtheorem{Assumption}{Assumption}

\usepackage{xargs}

\newcommand{\Cov}{{\rm Cov}}

\title{Causal Discovery in High-Dimensional Point Process Networks with Hidden Nodes
}
\author{Xu Wang and Ali Shojaie \\
Department of Biostatistics, University of Washington
}
\begin{document}

\maketitle

\begin{abstract}
Thanks to technological advances leading to near-continuous time observations, emerging multivariate point process data offer new opportunities for causal discovery. However, a key obstacle in achieving this goal is that many relevant processes may not be observed in practice. 
Na\"ive estimation approaches that ignore these hidden variables can generate misleading results because of the unadjusted confounding. To plug this gap, we propose a deconfounding procedure to estimate high-dimensional point process networks with only a subset of the nodes being observed. Our method allows flexible connections between the observed and unobserved processes. It also allows the number of unobserved processes to be unknown and potentially larger than the number of observed nodes. 
%With the confounding shrunk by trim transformation and further eliminated using penalized regression, 
% the trim transformation shrinks the size of confounding effects and then 
%
Theoretical analyses and numerical studies highlight the advantages of the proposed method in identifying causal interactions among the observed processes.
% We verify our method via extensive simulations and demonstrate its utility by analyzing neuronal spike train data.
\end{abstract}

\newpage 
%%%%%%%%%%%%%%%%%%%%%%%%%%%%%%%%%%%%%%%%%%
%%%%%%%%%%%%
\section{Introduction}\label{sec:intro}
%%%%%%%%%%%%
%\sw{1. Causal inf from time series is difficult for at least two reasons etc }
Learning causal interactions from observational multivariate time series is generally impossible. Among many challenges \citep{shojaie2021granger}, two of the most important ones are that i) the data acquisition rate may be much slower than the underlying
rate of changes; and ii) there may be unmeasured confounders \citep{Clark2019, Reid2019}.   
First, due to the cost or technological constraints, the data acquisition rate may be much slower than the underlying rate of changes. In such settings, the most commonly used procedure for inferring interactions among time series, Granger causality, may both miss true interactions and identify spurious ones \citep{BREITUNG2002, Silvestrini2008, tank2019}. 
Second, the available data may only include a small fraction of potentially relevant variables, leading to unmeasured confounders. 
%that can bias causal discovery.
%
Na\"ive connectivity estimators that ignore these confounding effects can produce highly biased results \citep{soudry2014shotgun}. 
Therefore, reliably distinguishing causal connections between pairs of observed processes from correlations induced by common inputs from unobserved confounders remains a key challenge. 

% \sw{
% 2. The first reason, subsampling comes up in many existing neuroscience data have this problem. However, new technologies, such as calcium florescent imaging that generate spike train data, make it possible to collect “live” data
% 3. The second issue, hidden nodes, is not as big of a deal with classical data (fMRI), but is a major issue with calcium florescent imaging and spike train data etc. However, because of 2), there is an opportunity for causal discovery if we can account for hidden nodes.
% }
% #

Learning causal interactions between neurons is critical to understanding cognitive functions. 
Many existing neuroscience data, such as data collected using functional magnetic resonance imaging (fMRI), have relatively low temporal resolutions, and are thus of limited utility for causal discovery \citep{Lin2014}. This is because many important neuronal processes and interactions happen at finer time scales \citep{Zhou2014AnalysisOS}.  
New technologies, such as calcium florescent imaging that generate spike train data, make it possible to collect ``live" data that are at high temporal resolutions \citep{Prevedel2014}.
The spike train data, which are  multivariate point processes containing spiking times of a collection of neurons, are increasingly used to learn the latent brain connectivity networks and to glean insight into how neurons respond to external stimuli \citep{Okatan2005}.
For example, \citet{Boldingeaat6904} collected spike train data on neurons in mouse olfactory bulb region at 30kHz under multiple laser intensity levels to study the odor identification mechanism. 
Despite progress in recording the  activity of massive populations of neurons \citep{Antal2013}, simultaneously monitoring a complete network of spiking neurons at high temporal resolutions is still beyond the reach of the current technology. 
In fact, most experiments only collect data on a small fraction of neurons, leaving many unobserved neurons \citep{trong_rieke_2008, Tchumatchenko2011, Huang2015}. 
These hidden neurons may potentially interact with the neurons inside the observed set and cannot be ignored. Nevertheless, given its high temporal resolution, spike train data provide an opportunity for causal discovery if we can account for the unmeasured confounders. 

% literature on existed methods 
% #4 There are some methods for dealing with hidden nodes, such as FCI etc, which may not work
% The problem of causal discovery in the presence of hidden variables has been studied using causal structural learning approaches, such as the Fast Causal Inference (FCI) algorithm and its variants \citep{Spirtes2000, Clark2019}. 
%
When unobserved confounders are a concern, causal effects among the observed variables can be learned using causal structural learning approaches, such as the Fast Causal Inference (FCI) algorithm and its variants \citep{Spirtes2000, Clark2019}. 
% Causal effects in the presence of hidden variables can be learned using causal structural learning approaches, such as the Fast Causal Inference (FCI) algorithm and its variants \citep{Spirtes2000, Clark2019}. 
However, these algorithms may not identify all causal edges. 
Specifically, instead of learning the directed acyclic graph (DAG) of causal interactions, FCI learns the maximally ancestral graph (MAG). This graph includes causal interactions between variables that are connected by directed edges, but also bi-directed edges among some other variables, leaving the corresponding causal relationships undetermined. 
As a result, causality discovery using these algorithms is not always satisfactory. For example, \citet{Malinsky2018} recently applied FCI to infer causal network of time series and found a low recall for identifying the true casual relationships.  
Additionally, despite recent efforts \citep{chen2021causal}, causal structure learning remains  computationally intensive, because the space of candidate causal graphs grows super-exponentially with the number of network nodes.

The Hawkes process \citep{Hawkes1971} is a popular model for analyzing multivariate point process data. In this model, the probability of future events for each component can depend on the entire history of events of other components. Under straightforward conditions, the multivariate Hawkes process reveals Granger causal interactions among multivariate point processes \citep{eichler2017graphical}. Moreover, assuming that all relevant processes are observed in a linear Hawkes process, causal interactions among components can also be inferred  \citep{Bacry2016}. 
The Hawkes process thus provides a flexible and interpretable framework for investigating the latent network of point processes and is widely used in neuroscience applications
\citep{Brillinger1988,Johnson1996,krumin_reutsky_shoham_2010, Pernice2011, Reynaud2013, TRUCCOLO2016336, LAMBERT20189}. 

In modern applications, it is common for the number of measured components, e.g., the number of neurons, to be large compared to the observed period, e.g., the duration of neuroscience experiments. The high-dimensional nature of data in such applications poses challenges to learning the connectivity network of a multivariate Hawkes process. To address this challenge, \citet{hansen2015} and \citet{Shizhe2017} proposed $\ell_1$-regularized estimation procedures and  \citet{wang2020statistical} recently developed a high-dimensional inference procedure to characterize the uncertainty of these regularized estimators.
%
% While assuming sufficient sampling is likely satisfied thanks to the high-frequency nature of the spike train data, observing the data from the entire neuron population is often impossible with the current technology. 
% %
% For example, in the experiment by \citet{Boldingeaat6904}, the spike train data are only available on a few neurons. % not sure if that is the best ref, since I talk high-dimension but later say hidden variables. 
%
However, due to the confounding from unobserved neurons in practice, existing estimation and inference procedures assuming complete observation from all components, may not provide reliable estimates.
%
%This is particularly the case in neuroscience applications, where discoveries on neural connectivies are limited to Granger causality, instead of true causality. % need reference

Accounting for unobserved confounders in high-dimensional regression has been the subject of recent research. Two such examples are HIVE \citep{bing2020adaptive} and trim regression \citep{cevid2020spectral}, which facilitate causal discovery using high-dimensional regression with unobserved confounders. 
However, these methods are designed for linear regression with independent observations and do not apply to the  long-history temporal dependency setting of Hawkes processes.
Moreover, they rely on specific assumptions on observed and unobsvered causal effects, which are not clear to hold in neuronal network settings. 

In this paper, we consider learning causal interactions among high-dimensional point processes with (potentially many) hidden confounders. Considering the generalization of the above two approaches to the setting of Hawkes processes, we show that the assumption required by trim regression is more likely to hold in a stable point process network, especially when the confounders affect many observed nodes. Motivated by this finding, we propose a generalization of the trim regression, termed \textit{hp-trim}, for causal discovery from high-dimensional point processes in the presence of (potentially many) hidden confounders. We establish a non-asymptotic convergence rate in estimating the network edges using this procedure. Unlike the previous result for independent data \citep{cevid2020spectral}, our result considers both the temporal dependence of the Hawkes processes as well as the network sparsity. 
Using simulated and real data, we also show that \textit{hp-trim} has superior finite-sample performance compared to the corresponding generalization of HIVE for point processes and/or the na\"ive approach that ignores the unobserved confounders. 

% 

%%%%%%%%%%%%%
\section{The Hawkes Processes with Unobserved Components} \label{sec:hawkes}
%%%%%%%%%%%%%
\subsection{The Hawkes Process}
Let $\{t_k\}_{k\in \mathbb{Z}}$ be a sequence of real-valued random variables,
taking values in $[0, T]$, with $t_{k+1} > t_k$ and $t_1 \ge 0$ almost surely.
Here, time $t = 0$ is a reference point in time, e.g., the start of an
experiment, and $T$ is the duration of the experiment. A simple point process
$N$ on $\mathbb{R}$ is defined as a family $\{ N(A) \}_{ A \in
\mathcal{B}(\mathbb{R}) }$, where $\mathcal{B}(\mathbb{R})$ denotes the Borel
$\sigma$-field of the real line and $N(A) = \sum_k \mathbf{1}_{\{t_k \in A\}} $.
The process $N$ is essentially a simple counting process with isolated jumps of
unit height that occur at $\{t_k\}_{k\in \mathbb{Z}}$. We write $ N([t, t + dt)
)$ as $dN(t)$, where $dt$ denotes an arbitrarily small increment of $t$.

Let $\mathbf{N}$ be a $p$-variate counting process $\mathbf{N} \equiv \{
N_i\}_{i\in \{1,\dots, p \}}$, where, as above, $N_i$ satisfies $N_i(A) = \sum_k
\mathbf{1}_{\{t_{ik} \in A\}}$ for $A \in \mathcal{B}(\mathbb{R})$ with $\{
t_{i1}, t_{i2}, \dots \}$ denoting the event times of $N_i$.  Let
$\mathcal{H}_t$ be the history of $\mathbf{N}$ prior to time $t$.  The intensity
process $\{ \lambda_{1}(t), \dots, \lambda_{p}(t) \}$ is a $p$-variate
$\mathcal{H}_t$-predictable process, defined as
\begin{align}\label{eq:intensity_function_prob}
	\lambda_i (t) dt & = \mathbb{P}(dN_{i}(t)=1 \mid \mathcal{H}_t )  .
\end{align}
\citet{Hawkes1971} proposed a class of point process models in which
past events can affect the probability of future events. The process $\mathbf{N}$ is a \textit{linear Hawkes process} if the intensity function for each unit $i \in \{1,\ldots,p\}$ takes the form
\begin{align}
  \lambda_{i}(t)
  &=   \mu_{i} + \sum_{j=1}^{p} \left(\omega_{ij} *  dN_{j} \right )(t)
    \label{eq:linear_hawkes}  ,
\end{align}
where
\begin{align}
  \left( \omega_{ij} *  dN_{j}  \right)(t)  =
  \int_0^{t-}  \omega_{ij}(t-s)   dN_{j} (s)
  = \sum_{ k: t_{jk} < t }   \omega_{ij}( t- t_{jk}) . \label{eq:transfer_function}
\end{align}
Here, $\mu_{i}$ is the background intensity of unit $i$ and
$\omega_{ij}(\cdot): \mathbb{R}^+ \rightarrow \mathbb{R}$ is the
\textit{transfer function}. In particular, $\omega_{ij}( t - t_{jk}) $
represents the influence from the $k$th event of unit $j$ on the intensity
of unit $i$ at time $t$.

Motivated by neuroscience applications \citep{Linderman2014, MAGRANSDEABRIL2018120}, we consider a parametric transfer function
$\omega_{ij}(\cdot)$ of the form
\begin{equation} \label{eq:omega}
\omega_{ij} (t)    =   \beta_{ij} \kappa_{j}(t) 
\end{equation}
with a \textit{transition kernel}
$\kappa_j(\cdot): \mathbb{R}^+ \rightarrow \mathbb{R}$ that captures the decay of the dependence on past events. This leads to $ \left( \omega_{ij} * dN_{j}  \right)(t) = \beta_{ij} x_{j}(t)$, where the \textit{integrated stochastic process}
\begin{equation}\label{eq:design_column_xt}
 x_{j}(t)  = \int_0^{t-} \kappa_{j}(t-s) dN_{j}(s) 
\end{equation}
summarizes the entire history of unit $j$ of the multivariate Hawkes processes. A commonly used example is the exponential transition kernel, $\kappa_{j}(t) = e^{-t}$ \citep{Bacry2015}.

Assuming that the model holds and all relevant processes are observed, it follows from \cite{Bacry2015} that the \textit{connectivity coefficient}  $\beta_{ij}$ represents the strength of the \textit{causal} dependence of unit $i$'s intensity on unit $j$'s past events. 
A positive $\beta_{ij}$ implies that past events of
unit $j$ \textit{excite} future events of unit $i$ and is often considered in the
literature \citep[see, e.g.,][]{Bacry2015, Etesami2016}. However, we might also wish to allow for negative $\beta_{ij}$ values to represent
\textit{inhibitory} effects 
\citep{Shizhe2017,costa2018}, which are expected in neuroscience applications \citep{Purves2001}. 

Denoting
$\bm{x}(t)= ( x_1(t), \dots, x_p(t) )^\top \in \mathbb{R}^{p}$ and
$\bm{\beta}_i = (\beta_{i1}, \dots, \beta_{ip})^\top \in
\mathbb{R}^{p}$, we can write
\begin{align}
  \lambda_{i}(t)
  &=   \mu_{i} + \bm{x}^\top(t)\bm{\beta}_i  \label{eq:linear_hawkes_para_transfer} .
\end{align}
Furthermore, let $Y_i(t) = {dN_i(t)}/{dt}$ and
$\epsilon_i(t) =Y_i(t) - \lambda_i(t) $.  Then the linear Hawkes
process can be written compactly as
\begin{align}
  Y_i(t) &= \mu_{i} + \bm{x}^\top(t)\bm{\beta}_i  + \epsilon_i(t) \label{eq:y_t}.
\end{align}

\subsection{The Confounded Hawkes Process}
Because of technology constraints, neuroscience experiments usually collect data from only a small portion of neurons. As a result, many other neurons that potentially interact with the observed neurons will be unobserved. 
Consider a network of $p+q$ counting processes, where we only observe the first $p$ components. The number of unobserved neurons, $q$, is usually unknown and likely much greater than $p$. 
Extending \eqref{eq:y_t} to include the unobserved components, 
we obtain the \textit{confounded Hawkes model},
% the intensity function in \eqref{eq:linear_hawkes_para_transfer}, we get 
% \begin{align}
% \label{eq:linear_hawkes_para_transfer_hidden} 
%   \lambda_{i}(t)
%   &=   \mu_{i} + \bm{x}^\top(t)\bm{\beta}_i 
%   + \bm{z}^\top (t) \bm{\delta}_i , 
% \end{align}
% and 
\begin{align}
\label{eq:hidden_var_model}
  Y_i(t) &= \mu_{i} + \bm{x}^\top(t)\bm{\beta}_i  + \bm{z}^\top (t) \bm{\delta}_i  + \epsilon_i(t) ,
\end{align}
in which $\bm{z}(t)= ( x_{p+1}(t), \dots, x_{p+q}(t) )^\top \in \mathbb{R}^{q}$ denotes the integrated processes of the hidden components, and $\bm{\delta}_i \in  \mathbb{R}^{q}$ denotes the connectivity coefficients from the unobserved components to unit $i$. 

\begin{figure}[t]
	\centering
	\includegraphics[width=1\linewidth, clip=TRUE, trim=0mm 0mm 0mm 0mm]{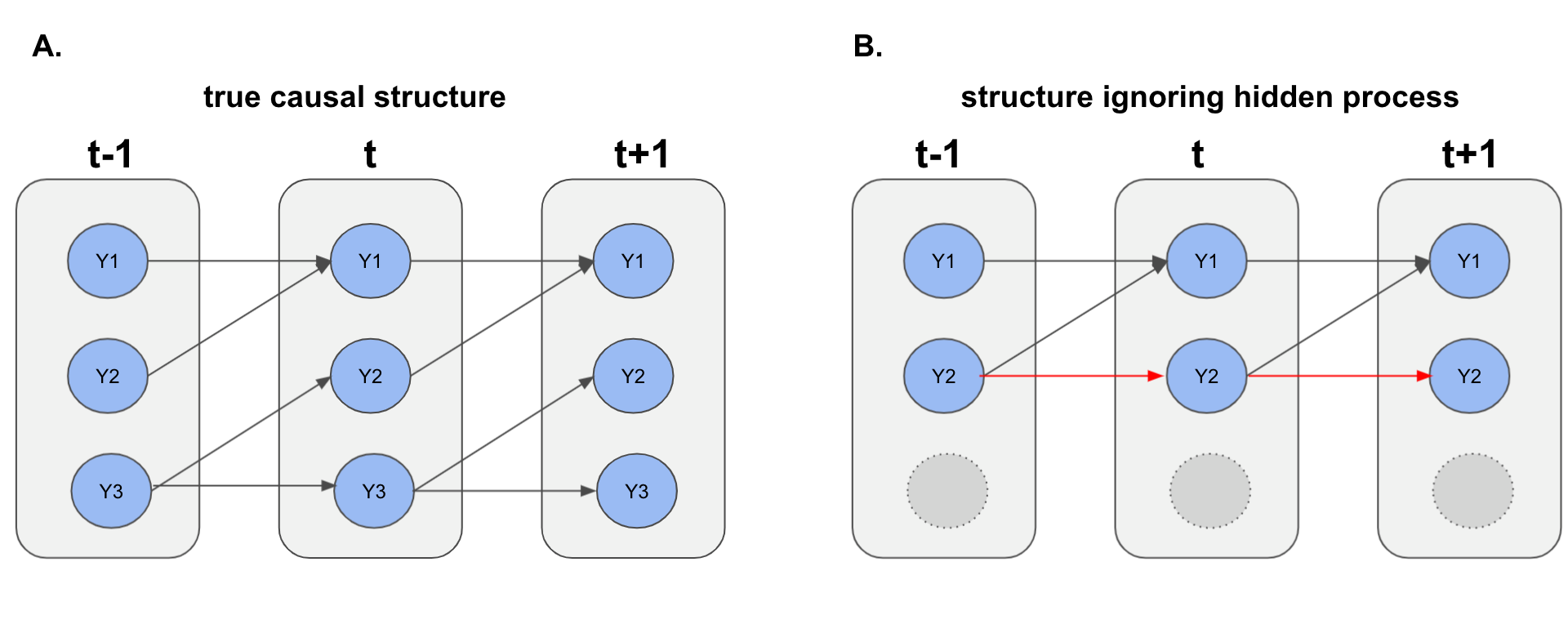}
	%\vspace{2mm}
	\caption{
	Illustration of the effect of hidden confounders on inferred causal interactions among the observed variables. A) The true causal diagram for the complete processes. B) The causal structure of the observed process when the hidden component, $Y_3$, is ignored, including a spurious autoregressive effect of $Y_2$ on its future values.  
% 	C) Structural equations illustrate induced spurious causal self-excitation effect of $x_2$ when $x_3$ is unobserved ($\epsilon_i^t, \epsilon_2^t,\epsilon_3^t$ are independent noises). 
	 }
		\label{fig:confounding_example_fig}
	\end{figure}

Unless the observed and unobserved processes are independent, the na\"ive estimator that ignores the unobserved components will produce misleading conclusion about the causal relationship among the observed components. This is illustrated in the simple linear vector autoregressive process of  Figure~\ref{fig:confounding_example_fig}. This example includes three continuous random variables generated according to the following set of equations 
%\as{it may be better to change the symbols in the figure and here to something other than x to avoid confusion with the integrated process}
\begin{align*}
    Y_1(t) &= Y_1(t-1) + Y_2(t-1) + \epsilon_1(t-1) \\
    Y_2(t) &= Y_3(t-1) + \epsilon_2(t-1) \\
    Y_3(t) &= Y_3(t-1) + \epsilon_2(t-1) ,
\end{align*}
where $\epsilon_i$ are mean zero innovation or error terms. 
The Granger causal network corresponding to the above process is shown in Figure~\ref{fig:confounding_example_fig}A. Figure~\ref{fig:confounding_example_fig}B shows that if $Y_3$ is not observed, the conditional means of the observed variables $Y_1$ and $Y_t$, namely, 
\begin{align*}
    \mathbb{E}\left\{Y_1(t) \mid Y_1(t-1), Y_2(t-1)\right\} &= Y_1(t-1) + Y_2(t-1) \\
    \mathbb{E}\left\{Y_2(t) \mid Y_1(t-1), Y_2(t-1)\right\} &= Y_2(t-1), 
\end{align*}
leads to incorrect Granger causal conclusions---in this case, a spurious autoregressive effect from the past values of $Y_2$. The same phenomenon occurs in Hawkes processes with unobserved components. 

% example structral equations to support the figure illustration 

% \[
%     \left\{\begin{aligned} 
%       x_1^t & = x_1(t-1) + x_2(t-1) + \epsilon_1^t \\
%     x_2^t & = x_3(t-1) + \epsilon_2^t \\
%     x_3^t &=  x_3(t-1) + \epsilon_3^t 
%     \end{aligned}\right. \Rightarrow
%     \left\{\begin{aligned}
%      \E \left \{ x^t_1 | x_1(t-1), x_2(t-1) \right \}  &=  x_1(t-1) + x_2(t-1) \\ 
%      \E \left \{ x^t_2 | x_1(t-1), x_2(t-1) \right \}  &=   \textcolor{red}{x_2(t-1) } 
%     \end{aligned}\right.
% \]

Throughout this paper, we assume that the confounded linear Hawkes model in \eqref{eq:hidden_var_model} is \textit{stationary}, meaning that for all units
$i=1,\dots, p$, the spontaneous rates $\mu_i$ and strengths of
transition $(\bm{\beta}_i, \bm{\delta}_i)$ are constant over the time range $[0,T]$
\citep{Bremaud1996, Daley2003}.

%%%%%%%%%%%%
% \section{The Proposed Methods}\label{sec:estimation}
\section{Estimating Causal Effects in Confounded Hawkes Processes} \label{sec:estimation}
%%%%%%%%%%%%
\subsection{Extending trim regression to Hawkes Processes}
% When the hidden component, $\bm{z}^\top(t)\bm{\delta}_i$, is correlated with the observed part, $\bm{x}^\top(t)\bm{\beta}_i$, directly estimating $\bm{\beta}_i$ ignoring the hidden variables generate biased results.
%
%
%
Let $\bm{b}_i \in \mathbb{R}^p$ be the projection coefficient of $ \bm{z}^\top (t) \bm{\delta}_i  $ onto $\bm{x}(t)$ such that 
\begin{align}
\label{eq:bi}
   \Cov\left(\bm{x}(t), \bm{z}^\top (t) \bm{\delta}_i - \bm{x}^\top(t)\bm{b}_i \right ) =0 . 
\end{align}
We can write the confounded linear Hawkes model in \eqref{eq:hidden_var_model} in the form of the \textit{perturbed linear model} \citep{cevid2020spectral}:
\begin{align}\label{eq:perturb}
  Y_i(t) &= \mu_{i} + \bm{x}^\top(t)\left( \bm{\beta}_i + \bm{b}_i \right) 
  + \nu_i(t), 
\end{align}
where $\nu_i(t)  = \left( \bm{z}^\top (t) \bm{\delta}_i - \bm{x}^\top(t) \bm{b}_i \right)+ \epsilon_i(t) $.
By the construction of $\bm{b}_i$, $\nu(t)$ is uncorrelated with the observed processes $\bm{x}(t)$ and
$\bm{b}_i $ represents the bias, or the perturbation, due to the confounding from $\bm{z}^\top (t) \bm{\delta}_i$. In general, $\bm{b}_i \ne  0$ unless $ \Cov( \bm{x}(t),  \bm{z}(t)) =0$.

The perturbed model in \eqref{eq:perturb} is generally unidentifiable because we can only estimate $ \bm{\beta}_i + \bm{b}_i  $ from the observed data, e.g., by regressing $Y_i(t)$ on $\bm{x}(t)$. 
The \textit{trim regression}  \citep{cevid2020spectral} is a two-step deconfounding procedure to estimate $\bm{\beta}_i$ for independent and Gaussian-distributed data. The method first applies a simple spectral transformation, called trim transformation (described below), to the observed data. It then estimates $\bm{\beta}_i$, using penalized regression.  
% We call this procedure \textit{trim regression} for short. 
When $\bm{b}_i$ is sufficiently small, the method consistently estimates $\bm{\beta}_i$. 
Although this condition is generally not valid for Gaussian-distributed data, previous work on Hawkes processes \citep{Shizhe2017} implies that the confounding magnitude cannot be large when the underlying network is stable, particularly when the confounders affect many observed components (see the discussion following Corollary~1 in Section~\ref{sec:theory}). This allows us to generalize the trim regression to learn the network of multivariate Hawkes processes. 

Assume, without loss of generality, that the first $p$ components are observed at times indexed from $1$ to $T$. Let $X \in \mathbb{R}^{T \times p}$ be the design matrix of the observed integrated process and $Y_i = \left( Y_i(1), \dots, Y_i(T) \right)^\top \in \mathbb{R}^T$ be the vector of observed outcomes. 
Further, let $X = UDV^\top$ be the singular value decomposition on $X$, where $U \in \mathbb{R}^{T\times r}$, $D \in \mathbb{R}^{r\times r}$ and $V \in \mathbb{R}^{p\times r}$; here, $r = \min(T, p)$ is the rank of $X$. Denoting the non-zero diagonal entries of $D$ by $d_1,\dots, d_r$, the \textit{spectral transformation} $F 
%\equiv  F(\tilde{d}_1,\dots, \tilde{d}_r ) 
: \mathbb{R}^{T \times p} \rightarrow \mathbb{R}^{T \times p } $ is given by
\begin{align}
    F = U \mat{ 
    \tilde{d_1}/d_1  & 0 & \dots & 0 \\
    0  & \tilde{d_2}/d_2 & \dots & 0 \\
    \vdots & \vdots  & \ddots & \vdots  \\
    0  & 0 & \dots & \tilde{d_r}/d_r \\
    } U^\top .
\end{align}
Denoting by $\widetilde{D}$ a diagonal matrix with entries $\tilde{d}_1,\dots, \tilde{d}_r$,  the first step of \textit{hp-trim} involves applying the spectral transformation to the observed data to obtain  
\begin{align}
    \widetilde{X} &= F X = U\widetilde{D} V^\top , \\
    \widetilde{Y} &= F Y  .
\end{align}
The spectral transformation is designed to reduce the magnitude of confounding. In particular, when $\bm{b}_i$ aligns with the top eigen-vectors of $X$, for an appropriate $F$, e.g., $\tilde{d}_k = \min( \tau, d_k ) $, the magnitude of $\widetilde{X}\bm{b}_i$ is small compared with $X\bm{b}_i$. Here, $\tau$ is a threshold parameter and the trim transformation is a special case of the spectral transformation when $\tau = \mathrm{median}\left( d_1, \dots, d_r \right)$. See \citet{cevid2020spectral} for additional details. 

In the second step, we then estimate the network connectivities using the transformed data by solving the following optimization problem
%using lasso \citep{TibsLasso1996} for $i\in \left\{1,\dots, p\right\}$:
\begin{align} \label{eq:lasso}
%   \widehat{\mu}_i , \widehat{\bm{\beta}}_i 
%     =  
\arg \min_{
	\substack{
	\mu_i \in \mathbb{R}, \bm{\beta}_i 
		\in \mathbb{R}^p\\ 1\le i \le p  
	} 
}
\sum_{i=1}^p 
\left\{ 
    \frac{1}{T}  
    \left \lVert \widetilde{Y}_i - \mu_i - \widetilde{X}\bm{\beta}_i \right \rVert_2^2+ \lambda \left \lVert \bm{\beta}_i \right \rVert_1 
\right\}
    ,
\end{align}
which is an instance of lasso regression \citep{TibsLasso1996} and can be solved separately for each $i \in\{ 1,\dots, p\}$.

\subsection{An alternative approach} 
% concise describe HIVE method steps and refer to appendix 
HIdden Variable adjustment Estimation (HIVE) \citep{bing2020adaptive} is an alternative method for estimating coefficients of a linear model with independent and Gaussian-distributed data in the presence of latent variables. Adapted to the network of multivariate point processes, HIVE first estimates the latent column space of the unobserved connectivity matrix, 
% %
% $\Delta = \mat{ \bm{\delta}^\top_1 \\ \dots \\ \bm{\delta}^\top_p } \in \mathbb{R}^{p \times q}$, 
% %
%
%\as{
$\Delta = \mat{ \bm{\delta}_1 & \dots&   \bm{\delta}_p  } ^\top \in \mathbf{R}^{p \times q} $
%} %\sw{this is correct}
%\mat{ \bm{\delta}^\top_1 \\ \dots \\ \bm{\delta}^\top_p } \in \mathbf{R}^{p \times q} $
, 
with $\bm{\delta}_i$ defined in \eqref{eq:hidden_var_model}. 
It then projects the outcome vector, $Y(t) = \left(  Y_1(t), \dots, Y_p(t) \right)^\top$, onto %\sw{the orthogonal space to the column space of $\Delta$}. 
the space orthogonal to the column space of $\Delta$. 
Assuming that the column space of the observed connectivity matrix, %$\Theta = \mat{  \bm{\beta}^\top_1 \\ \dots \\  \bm{\beta}^\top_p } \in \mathbb{R}^{p \times p}$, 
$
%\as{
\Theta = \mat{  \bm{\beta}_1 & \dots &  \bm{\beta}_p }^\top \in \mathbf{R}^{p \times p}
%}
$ %\sw{this is correct}
is orthogonal to that of $\Delta$, HIVE consistently estimates $\Theta $ using the transformed data.
While the orthogonality assumption might be satisfied when the hidden processes are external, such as experimental perturbations in genetic studies \citep{shawn2017}, it might be too stringent in a network setting. However, when the orthogonality assumption fails, HIVE may lead to poor edge selection performance, and potentially worse than the na\"ive method that ignores the hidden processes. HIVE also requires the number of hidden variables to be known. Although methods in selecting the number of hidden variables have been proposed, the resulting theoretical guarantees would only be asymptotic. An over- or under-estimated number can either miss the true edges or generate false ones. Given these limitations, we outline the extension of HIVE for Hawkes processes in Appendix~\ref{sec:hive} and refer the interested reader to \citet{bing2020adaptive} for details.
%

%%%%%%%%%%%%
\section{Theoretical Properties}\label{sec:theory}
%%%%%%%%%%%%
%
In this section we establish the recovery of the network connectivity in the presence of hidden processes. Technical proofs for the results in this section are given in Appendix~\ref{sec:proofs}.

We start by stating our assumptions. 
For a square matrix $A$, let $\Lambda_{\max}(A)$ and $\Lambda_{\min}(A)$ be its maximum and minimum eigenvalues, respectively. 

\begin{Assumption}\label{assumption1} Let $\Omega = \{ \Omega_{ij} \}_{1\le i,j
\le p+q} \in \mathbb{R}^{(p+q)\times (p+q)}$ with entries $\Omega_{ij} =   \int_0^{\infty}
|\omega_{ij}(\Delta)| d\Delta$. There exists a constant $\gamma_{\Omega}$ such
that $\Lambda_{\max} ( \Omega^T \Omega ) \le \gamma^2_{\Omega}  < 1 $.
\end{Assumption}

Assumption~\ref{assumption1} is necessary for stationarity of a Hawkes process \citep{Shizhe2017}. The constant $\gamma_{\Omega}$ does not depend on the
dimension $p+q$. For any fixed dimension, \citet{Bremaud1996} show that given this assumption the intensity process of the form \eqref{eq:linear_hawkes_para_transfer} is stable
in distribution and, thus, a stationary process exists. 
Since our connectivity coefficients of interest are ill-defined without stationarity, this assumption provides the necessary context for our estimation framework.

\begin{Assumption}\label{assumption2}
	There exists $\lambda_{\min}$ and $\lambda_{\max}$ such that
	$$
	0 < \lambda_{\min} \le \lambda_{i}(t) \le \lambda_{\max} < \infty ,
	\quad t\in [0, T]
	$$
	for all $i=1,\dots, p+q$.
\end{Assumption}
Assumption~\ref{assumption2} requires that the intensity rate is strictly
bounded, which prevents degenerate processes for all components of
the multivariate Hawkes processes. This assumption has been considered in the previous analysis of Hawkes processes \citep{ hansen2015,costa2018,Shizhe2017, wang2020statistical, cai2020latent}. %\as{add other refs in JE}

\begin{Assumption}\label{assumption3}
	The transition kernel $\kappa_j(t)$ is bounded and integrable over $[0,T]$, for $1\le j \le p+q$.
\end{Assumption}
%
% Assumption~\ref{assumption3} implies that the integrated process $x_j(t)$
% in \eqref{eq:design_column_xt} is bounded. 

\begin{Assumption}\label{assumption4}
	There exists constants $\rho_r \in (0,1)$ and $0< \rho_c < \infty $ such that
	\begin{align*}
	\max_{1\le i \le p+q}  \sum_{j=1}^{p+q} \Omega_{ij} \le \rho_r
	\qquad\text{and}\qquad
	\max_{1\le j \le p+q}   \sum_{i=1}^{p+q} \Omega_{ij} \le \rho_c .
	\end{align*}
\end{Assumption}
%
% we use this assumption to valid that the confounding size can be small, so we can not relax this assumption using the small confounding size.
%
Assumption~\ref{assumption3} implies that the integrated process $x_j(t)$
in \eqref{eq:design_column_xt} is bounded. 
Assumption~\ref{assumption4} requires maximum in- and out- intensity flows to be bounded, which provides a sufficient condition for bounding the eigenvalues of the cross-covariance of $\bm{x}(t)$ \citep{wang2020statistical}. A similar assumption is considered by \citet{Basu2015} in the context of VAR models. 
Together, Assumptions~\ref{assumption3} and \ref{assumption4} imply that the model parameters
are bounded, which is often required in time-series analysis \citep{Safikhani2017JointSB}. 
Specifically, these assumptions restrict the influence of the hidden processes from being too large. 

Define the set of active indices among the observed components, $S_i= \{j:  \beta_{ij}\ne 0 , 1\le j \le p \}$, and $s_i = |S_i|$ and $s^* \equiv \max_{ 1\le i \le p} s_i $. Let $Q = \frac{1}{T}   \sum_{t=1}^T   \mat{1 \\ \bm{x}(t) } 
\mat{1 &  \bm{x}^\top(t)}   $, and $\gamma_{\min} \equiv \Lambda_{\min} \left( Q \right)$ and $\gamma_{\max} \equiv \Lambda_{\max}\left( Q \right)$. Our first result provides a fixed sample bound on the error of estimating the connectivity coefficients. 
%
% Let 
% $
% \mathcal{C} =
% \left \{ 
% \Delta \in R^{p+1} : 
% \lVert \Delta_{S^c}\rVert_1 
% \le 3 \lVert \Delta_S \rVert_1 \right \}.
% $ 

\begin{Theorem}
	\label{theorem1}
	Suppose each of the $p$-variate Hawkes processes with intensity function defined in \eqref{eq:hidden_var_model}  satisfies Assumptions~\ref{assumption1}--~\ref{assumption4}. 
%	In addition, Conditions~\ref{def:RSC} and~\ref{def:tail_bound} are met.
	Assume $\log p \vee   (s^*)^{1/2} = o(T^{1/5})$. Then, taking $\lambda 
	= O(\Lambda^2_{\max}\left( F\right) T^{-2/5})$,
	\begin{align*}
	\left \lVert \bm{\beta}_i - \widehat{\bm{\beta}}_i \right \rVert_1 \le 
	C_1  \Lambda^2_{\max}(F) \frac{s^*}{ \gamma^2_{\min} } T^{-2/5}
+     
  C_2  \Lambda^{-2}_{\max}(F) T^{-3/5}\left \lVert  \widetilde{X}\bm{b}_i \right \rVert^2_2 ,
	\quad 1\le i \le p,
	\end{align*}
	with probability at least $1-c_1 p^2 T \exp(- c_2 T^{1/5})$, where $C_1, C_2, c_1, c_2 >0$ depend on the model parameters and the transition kernel. 
\end{Theorem}
%
%While the skeleton of our proof is similar to \citet[][Theorem~2]{cevid2020spectral}, we consider Hawkes process data that are defined on a continuous time domain. 
Compared to the case with independent and Gaussian-distributed data \citep[][Theorem~2]{cevid2020spectral}, we obtain a slower convergence rate because of the complex dependency of the Hawkes processes. Our rate takes into account the network sparsity among the observed components. It also does not depend on the size of unobserved components, $q$, which is critical in neuroscience experiments because $q$ is often unknown and potentially very large.  

The result in Theorem~\ref{theorem1} is different from the corresponding result obtained when all processes are observed \citep[][Lemma~10]{wang2020statistical}. More specifically, our result includes an extra error term, $\lVert \widetilde{X} \bm{b}_i \rVert^2_2$, which captures the effect of unobserved processes. Next, we show that when $\lVert \bm{b}_i  \rVert^2_2$ is sufficiently small, we obtain a similar rate of convergence as the one obtained when all processes are observed. 

% While trim transformation empirically makes this term small, the convergence of our estimator does not depend on such empirical behavior because $\Lambda_{\max}(F) \in (\gamma_{\min}, \gamma_{\max})$.

\begin{Corollary}
\label{corollary1}
Under the same assumptions in Theorem~\ref{theorem1}, suppose, in addition, $\lVert \bm{b}_i  \rVert^2_2 = O\left( \frac{s^*}{\gamma^2_{\min} \gamma_{\max} } T^{-4/5} \Lambda_{\max}^2(F) \right)$, 
	\begin{align*}
    \left \lVert \bm{\beta}_i - \widehat{\bm{\beta}}_i \right \rVert_1
	= O\left(  \frac{s^*}{\gamma^2_{\min} } \Lambda^2_{\max}\left(F\right) T^{-2/5} 	\right) ,
	\quad 1\le i \le p,
	\end{align*}
with probability at least $1-c_1 p^2 T \exp(- c_2 T^{1/5})$, where $c_1, c_2 >0$ depending on the model parameters and the transition kernel. 
\end{Corollary}

The spectral transformation empirically reduces the magnitude of $\frac{1}{T} \lVert \widetilde{X} \bm{b}_i \rVert^2_2$, especially when the confounding vector, $\bm{b}_i$, stays in the sub-space spanned by top right singular vectors of $X$; however, this is not guaranteed to hold for arbitrary $\bm{b}_i$. Corollary~\ref{corollary1} specifies a condition on $\bm{b}_i$ that leads to consistent estimation of $\bm{\beta}_i$, regardless of the empirical performance of the spectral transformation. 
While the condition does not always hold for arbitrary stochastic process, it is satisfied for a stable network of high-dimensional multivariate Hawkes processes when the confounding is dense.
Specifically, by the construction of $\bm{b}_i$ in \eqref{eq:bi}, Assumption~\ref{assumption4} implies that $ \lVert \bm{b}_i \rVert_1   
%=  \lVert \Var^{-1}( \bm{x}(t) ) \Cov(\bm{x}(t), \bm{z}(t))   \bm{\delta}_i  \rVert_1 
= O\left( \lVert  \bm{\delta}_i  \rVert_1 \right) 
= O(1)$.
% | \Var( \bm{x}(t) )  b|_1 \le  | \Cov(\bm{x}(t), \bm{z}(t))|_\infty  |delta|_1
% |\Cov(\bm{x}(t), \bm{z}(t))|_\infty = max row sum 
% bounded by row sum of transition matrix in Assumption 1-4 based on the Wiener-Hopf integral equations following a similar argument as Shizhe's EJS paper Theorem~1. 
% Then, | \Var( \bm{x}(t) )  b|_1  = O(1) 
% assuming |b|_0 = O(p), for those b_i ne 0, let A = \Var( \bm{x}(t) ), there exits c_i > 0, s.t., 
%   |sum_j a_ij  b_j| > c_i |a_ii*b_i| since min_eigen of A > 0, so no b s.t. Ab = 0.
% therefore, | \Var( \bm{x}(t) )  b|_1  > min (c_i*a_ii) |b|_1 
% thus, |b|_1 = O(1), which gives O(|b|^2_2)= O(1/p). 
When the confounding effects are relatively dense---i.e., $\lVert \bm{b}_i \rVert_0 = O(p)$, meaning that there are large number of interactions from unobserved nodes to the observed ones---we obtain $\lVert \bm{b}_i  \rVert^2_2  = O(1/p)$.
%Given the observed neuron population is only a tiny part of the entire neurons, the dense confounding condition likely happens. 
Therefore, the constraint on $\lVert \bm{b}_i  \rVert^2_2$ is likely satisfied under a high-dimensional network, when $p\gg T$. The high-dimensional network setting is common in modern neuroscience experiments where the number of neurons is often large compared to the duration of experiments.   

Next we introduce an additional assumption to establish the edge selection consistency. To this end, we consider the \textit{thresholded connectivity estimator}, 
$$ 
\widetilde{\beta}_{ij}  = \widehat{\beta}_{ij} \mathbf{1} \left( \left |\widehat{\beta}_{ij} \right | >  \tau \right ),  \quad 1\le i,j \le p .$$ 
Thresholded estimators are used for variable selections in high-dimensional network estimation \citep{shojaie2012adaptive} as they alleviate the need for restrictive irrepresentability assumptions  \citep{vandeGeer2011EJS}. 

\begin{Assumption}
	\label{assumption5}
	There exists $\tau >0 $ such that
	\begin{align*}
	\min_{ 1\le i,j \le p} \beta_{ij} \ge \beta_{min} > 2\tau .
	\end{align*}
\end{Assumption}
Assumption~\ref{assumption5} is called the $\beta$-$\min$ condition \citep{buhlmann2013} and requires sufficient signal strength for the true edges in order to distinguish them from $0$. %
% We construct the \textit{thresholded connectivity estimator} 
% $$ 
% \widetilde{\beta}_{ij}  = \widehat{\beta}_{ij} \mathbf{1} \left( \left |\widehat{\beta}_{ij} \right | >  \tau \right ),  \quad 1\le i,j \le p .$$ 
% Such thresholded estimator is often used for variable selections in high dimensions \citep{vandeGeer2011EJS}. 
Let the estimated edge set $\widehat{S} =\left  \{ (i,j): \widetilde{\beta}_{ij}  \ne 0 , 1\le i,j \le p \right \}$ and the true edge set
$S =\left  \{ (i,j): \beta_{ij}  \ne 0 , 1\le i,j \le p \right \}$.
The next result shows that the estimated edge set consistently recovers the true edge set. %in the presence of unobserved confounders.

\begin{Theorem}
	\label{theorem2}
Under the same conditions in Theorem~\ref{theorem1}, assume Assumption~\ref{assumption5} is satisfied with $\tau = O\left(  \frac{s^*}{\gamma^2_{\min} } \Lambda_{\max}^2(F) T^{-2/5} \right)$. Then,  
	\begin{align*}
	\mathbb{P}\left(   \widehat{S}  = S  \right )
	\ge 1-c_1 p^2 T \exp\left(- c_2 T^{1/5}\right) ,
	\end{align*}	
where $c_1, c_2 >0$ depending on the model parameters and the transition kernel. 
\end{Theorem}

Theorem~\ref{theorem2} guarantees the recovery of causal interactions among the observed components. As before, the result is valid irrespsective of the number of unobserved components, which is important in neuroscience applications. 

% We add the proofs of both theorems and the corollary in Appendix~\ref{sec:proofs}.

%%%

%%%%%%%%%%%%
\section{Simulation Studies}\label{sec:sim}
%%%%%%%%%%%%
We compare our proposed method, hp-trim, with two alternatives, HIVE and the na\"ive approach that ignores the unobserved nodes. To this end, we compare the methods in terms of their abilities to identify the correct causal interactions among the observed components.  

\begin{figure}%
    \centering
    \subfloat[\centering]{{\includegraphics[width=0.4\linewidth]{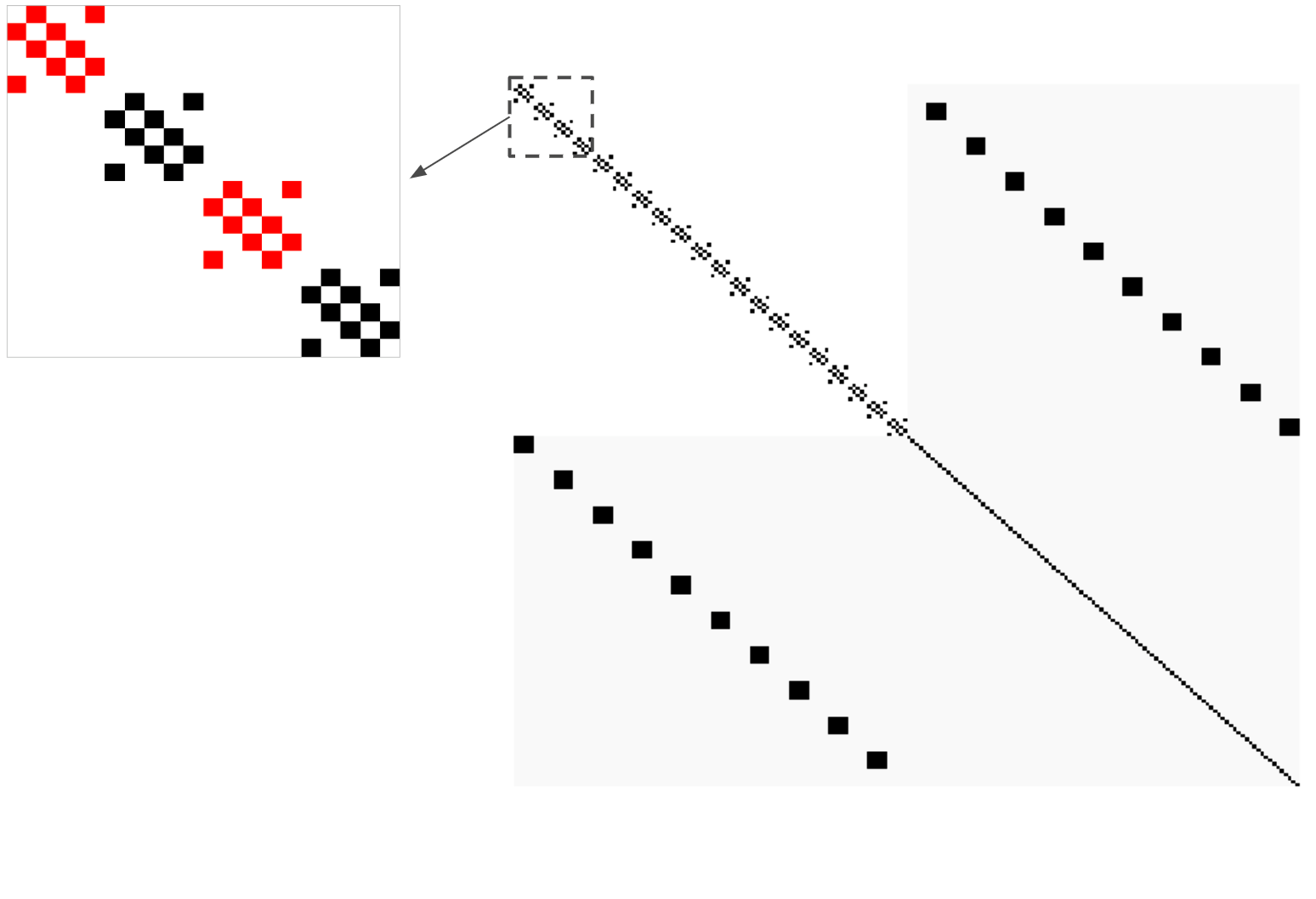} }
    }%
    \qquad
    \quad
    \subfloat[\centering ]{{\includegraphics[width=0.45\linewidth]{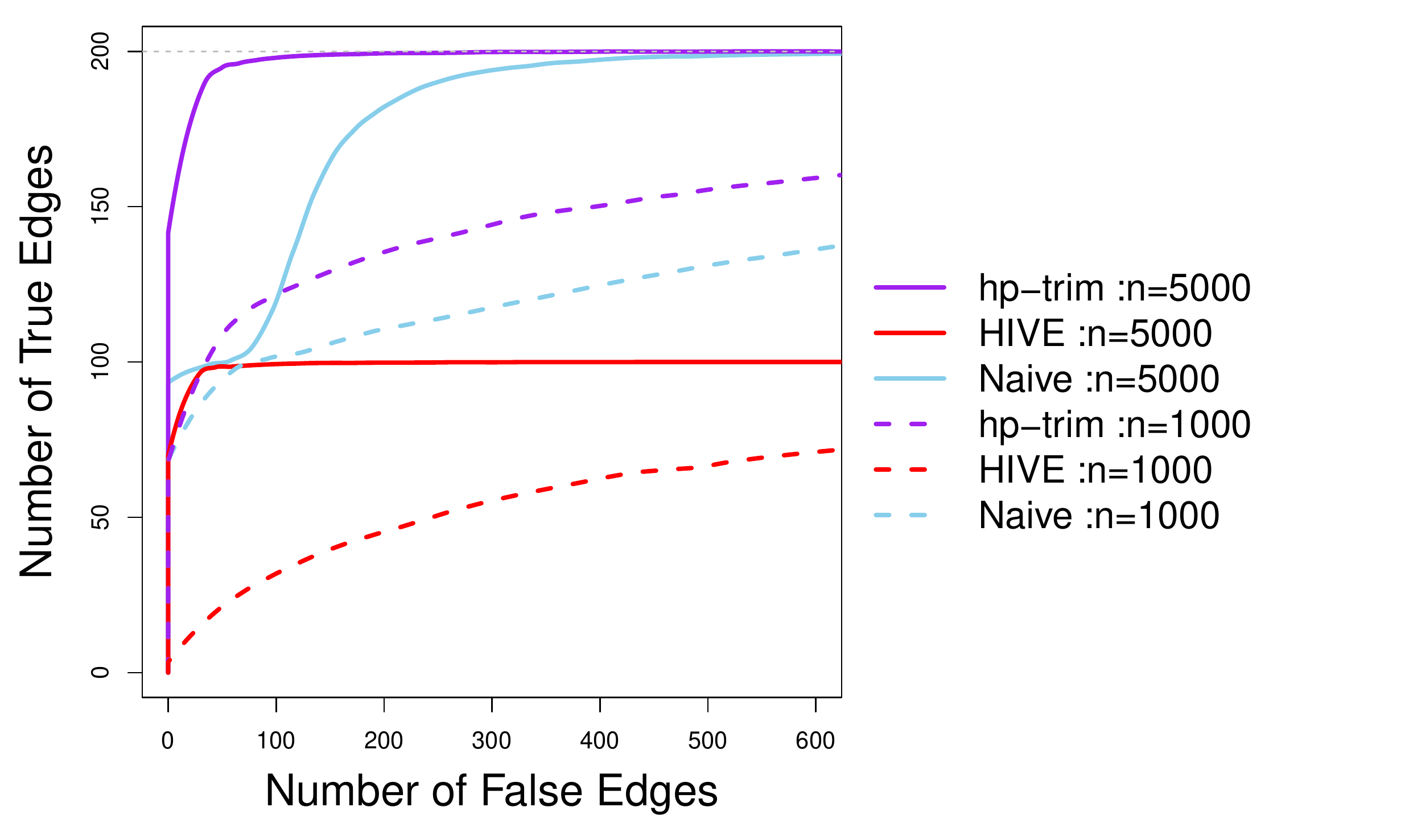} }}%
  %  \subfloat[\centering ]{{\includegraphics[width=0.5\linewidth]{Unorth_allline.pdf} }}%
    \caption{Edge selection performance of the proposed hp-trim approach compared with estimators based on HIVE (run with the known (oracle) number of latent features) and the na\"ive approach. Here, $p = q = 100$. (a) Visualization of the connectivity matrix, with unobserved connecitivies colored in gray and entries corresponding to edges shown in black.  
    This setting violates the orthogonality condition of HIVE because of the connections between the observed and the hidden nodes (represented by the non-zero coefficients colored in red). 
    (b) Average number of true positive and false positive edges detected using each method over 100 simulation runs.  }%
    \label{fig:unorth}%
\end{figure}

\begin{figure}%
    \centering
    \subfloat[\centering]{{\includegraphics[width=0.4\linewidth]{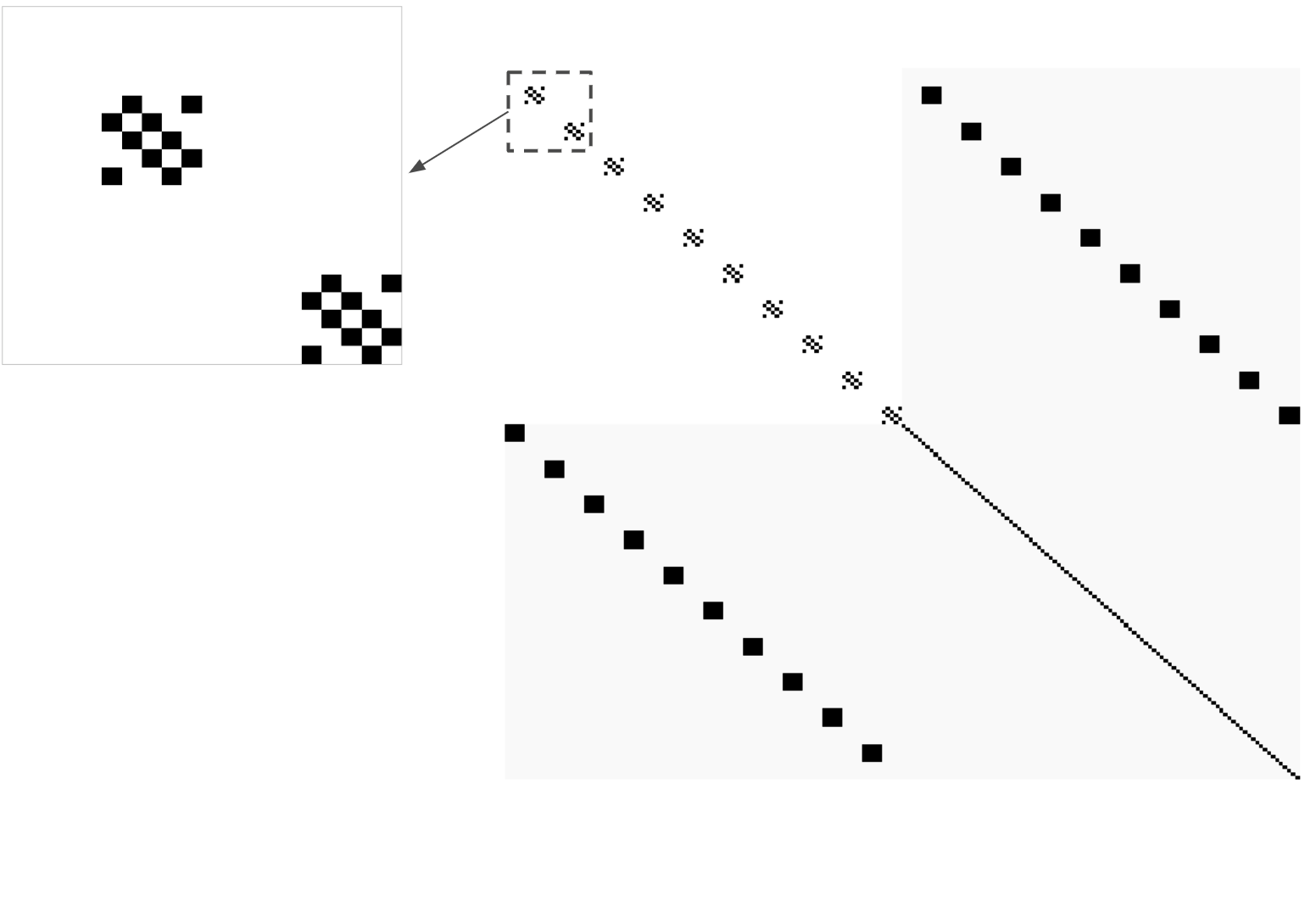} }}%
    \qquad
    \subfloat[\centering ]{{\includegraphics[width=0.45\linewidth]{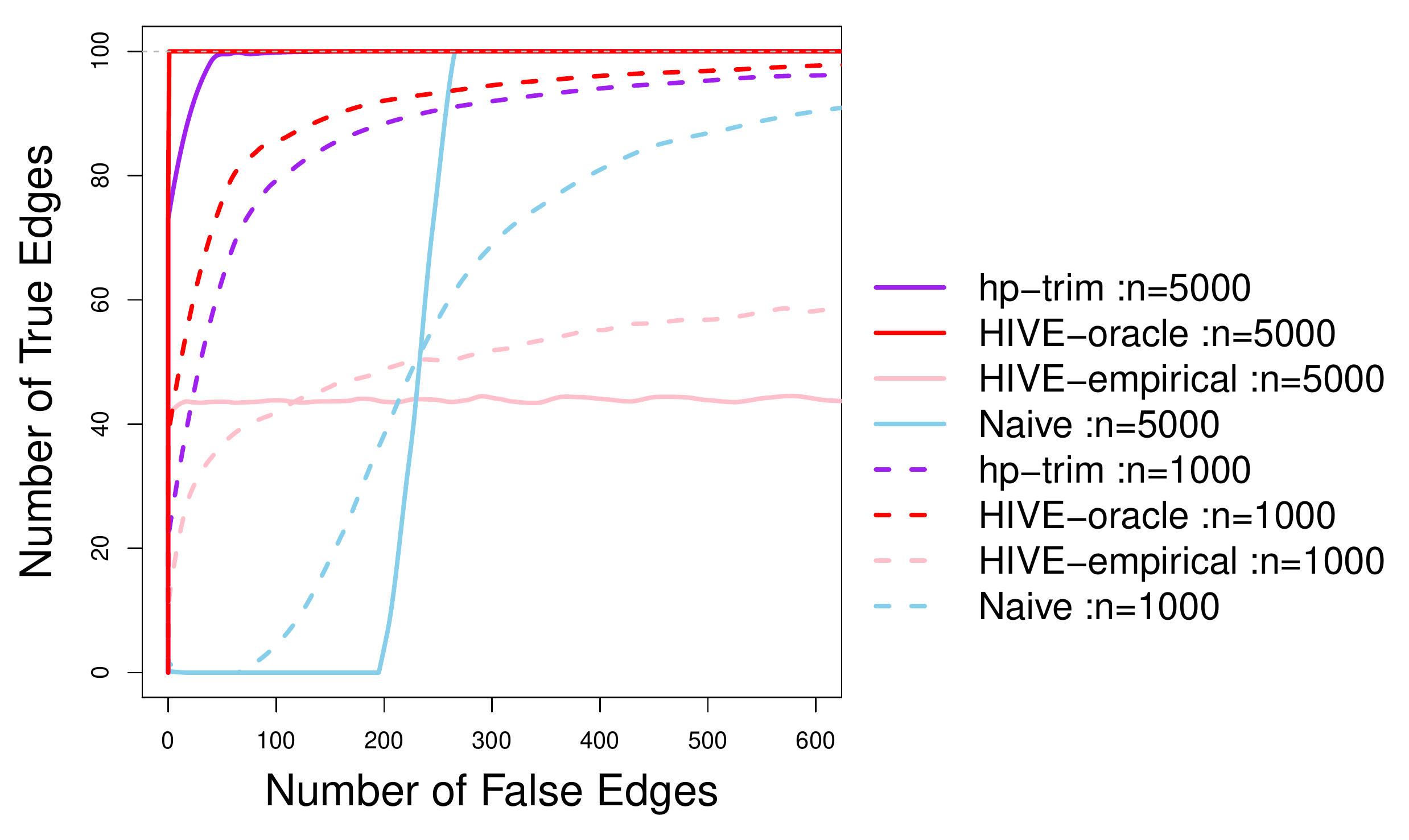} }}%
    %\subfloat[\centering ]{{\includegraphics[width=0.5\linewidth]{Orth_allline.pdf} }}%
   %\subfloat[\centering ]{{\includegraphics[width=0.5\linewidth]{Orth_allempircalq.pdf} }}%
    \caption{Edge selection performance of the proposed hp-trim approach compared with estimators based on HIVE and the na\"ive approach. Here, $p = q = 100$. (a) Visualization of the connectivity matrix, with unobserved connecitivies colored in gray and entries corresponding to edges shown in black. This setting satisfies the orthogonality condition of HIVE, which is run both with and without assuming known number of latent features. These two versions are denoted HIVE-oracle and HIVE-empirical, respectively. In HIVE-empirical the number of latent factors is estimated based on the estimate with highest frequency over the 100 simulation runs (estimated $\hat q=79$).  
    (b) Average number of true positive and false positive edges detected using each method over 100 simulation runs. %
    }%
    \label{fig:orth}%
\end{figure}

We consider a point process network consisting of $200$ nodes with half of the nodes being observed; that is $p = q = 100$. The observed nodes are connected in blocks of five nodes, and half of the blocks are connected with the unobserved nodes (see~Figure~\ref{fig:unorth}a). This setting exemplifies neuroscience applications, where the orthogonality assumption of HIVE is violated. 
As a sensitivity analysis, we also consider a second setting similar to the first, in which we remove the connections of the blocks that are not connected with the unobserved nodes %
This setting, shown in Figure~\ref{fig:orth}a, satisfies HIVE's orthogonality assumption. 

To generate point process data, we consider $\beta_{ij} = 0.12$ and $\delta_{ij} = 0.10$ in the setting of Figure~\ref{fig:unorth}a, and $\beta_{ij} = 0.2$ and $\delta_{ij} = 0.18$ in the setting of Figure~\ref{fig:orth}b. The background intensity, $\mu_i$, is set to $0.05$ in both settings. The transfer kernel function is chosen to be $\exp(-t)$. These settings satisfy the assumptions of stationary Hawkes processes. In both settings, we set the length of the time series to $T \in \{ 1000, 5000\}$ . 

The results in Figure~\ref{fig:unorth}b shown that hp-trim offers superior performance for both small and large sample sizes in the first setting. HIVE performs poorly, worse than the na\"ive approach, because the orthogonality condition is violated. When the orthogonality condition is satisfied (Figure~\ref{fig:orth}b), HIVE shows the best performance. However, this advantage requires knowledge of the correct number of latent features. When the number of latent features is unknown and estimated from data, HIVE's performance deteriorates, especially with an insufficient sample size. In contrast, hp-trim's performance with both moderate and large sample sizes is close to the oracle version of HIVE (HIVE-oracle).

%%%%%%%%%%%%%%%%%
\section{Analysis of Mouse Spike Train Data}\label{sec:data}

We consider the task of learning causal interactions among the observed population of neurons, using the spike train data from \citet{Boldingeaat6904}.
In this experiment, spike times are recorded at 30 kHz on a region of the mice olfactory bulb (OB), while a laser pulse is applied directly on the OB cells of the subject mouse. The laser pulse has been applied at increasing intensities from 0 to 50 ($mW/mm^2$). The laser pulse at each intensity level lasts 10 seconds and is repeated 10 times on the same set of neuron cells of the subject mouse.

The experiment consists of spike train data multiple mice and we consider data from the subject mouse with the most detected neurons ($25$) under laser (20 $mW/mm^2$) and no laser conditions. In particular, we use the spike train data from one laser pulse at each intensity level. Since one laser pulse spans 10 seconds and the spike train data is recorded at 30~kHz, there are 300,000 time points per experimental replicate. 

\begin{figure}[!t]
	\centering
	\includegraphics[width=1\linewidth, clip=TRUE, trim=0mm 0mm 0mm 0mm]{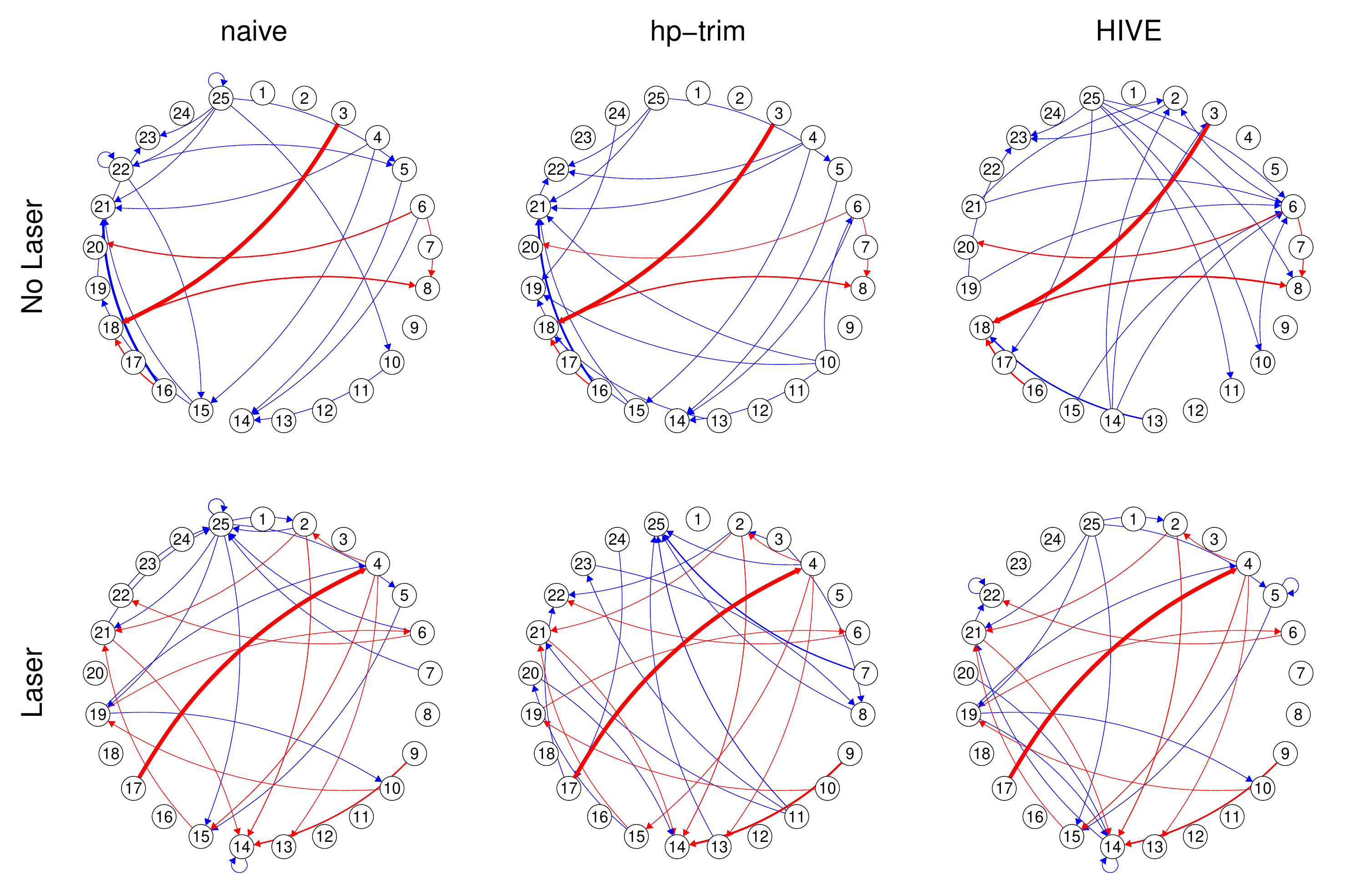}
	\caption{Estimated functional connectivities among neurons using mouse spike train data from laser and no-laser conditions \citep{Boldingeaat6904}.
	Common edges estimated by the three methods are in red and the method-specific edges are in blue. Thicker edges indicate estimated connectivity coefficients of larger magnitudes. %\as{change trim to hp-trim here and elsewhere (if you think that is a good name)}
	}
		\label{fig:dataexample}
	\end{figure}

The population of observed neurons is a small subset of all the neurons in mouse's brain. Therefore, to discover causal interactions among the $p=25$ observed neurons, we apply our estimation procedure, hp-trim, along with HIVE and na\"ive approaches, separately for each intensity level, and obtain the estimated connectivity coefficients for the observed neurons. %
For ease of comparison, the tuning parameters for both methods are chosen to have about 30 estimated edges; 
moreover, for HIVE, $q$ is estimated following the procedure in \citet{bing2020adaptive}, which is based on the maximum decrease in eigenvalue of the covariance matrix of the errors, $\widetilde{E}(t)$ in \eqref{eq:hive_model}.

Figure~\ref{fig:dataexample} shows the estimated connectivity coefficients specific to each laser condition in a graph representation. In this representation, each node represents a neuron, and a directed edge indicates a non-zero estimated connectivity coefficient. We see different network connectivity structures when laser stimulus is applied, which agrees with the observation by neuroscientists that the OB response is sensitive to the external stimuli \citep{Boldingeaat6904}. 
Compared to our proposed method, the na\"ive approach generates a more similar network than HIVE under both laser and no-laser conditions, which is likely an indication that the na\"ive estimate is incorrect in this application.

\begin{figure}[!t]
	\centering
	\includegraphics[width=.4\linewidth, clip=TRUE, trim=0mm 0mm 0mm 0mm]
	{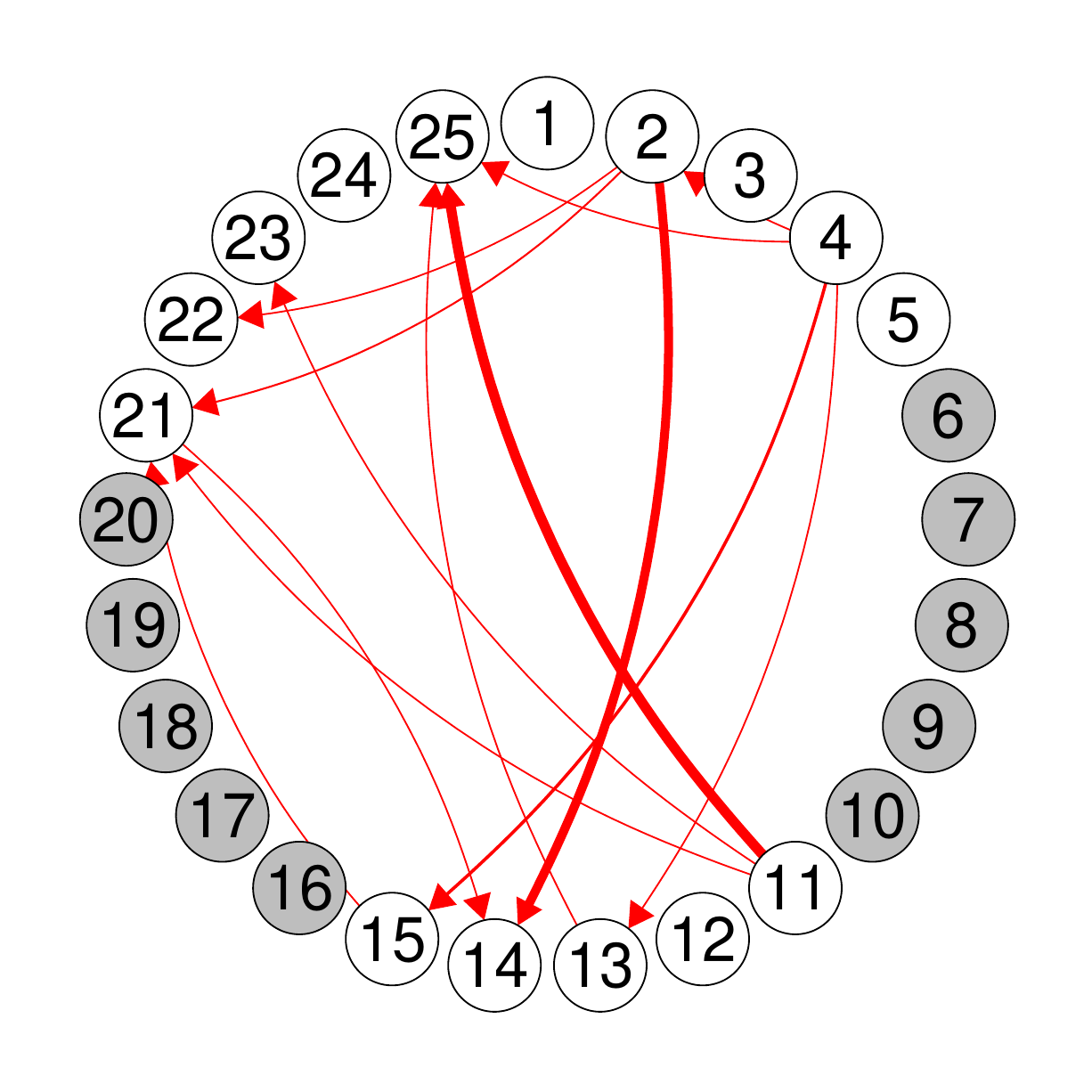}
	\caption{Estimated functional connectivities using hp-trim among a subset of neurons. Here, data is the same as that used in Figure~\ref{fig:dataexample}, except that $10$ neurons (shown in gray) are considered hidden. 
	Thicker edges indicate estimated connectivity coefficients of larger magnitudes. All estimated edges using the subset of neurons are also found in the estimated network using all neurons in Figure~\ref{fig:dataexample}. 
	 }
		\label{fig:dataexample_subset}
	\end{figure}

%\textcolor{orange}{
As discussed in Section~\ref{sec:theory}, our inference procedure is asymptotically valid. In other words, with large enough sample size, if the other assumptions in Section~\ref{sec:theory} are satisfied, the estimated edges should represent the true edges. Assessing the validity of the assumptions and selecting the true edges in real data applications is challenging. However, we can assess the sample size requirement and the validity of assumptions by estimating the edges over a subset of neurons as if the other removed neurons are unobserved. If the sample size is sufficient and the other assumptions are satisfied, we should obtain similar connectivities among the observed subset of neurons, even when some neurons are hidden. 
Figure~\ref{fig:dataexample_subset} shows the result of such a stability analysis for the laser condition using hp-trim. Comparing the connectivities in this graph with those in Figure~\ref{fig:dataexample} indicates that the estimated edges using the subset of neurons are consistent with those estimated using all neurons. Thus, the assumptions are likely satisfied in this application.
%}

%%%%%%%%%%%%
\section{Discussion}\label{sec:discussion}
%%%%%%%%%%%%

We proposed a causal-estimation procedure with theoretical guarantees for high-dimensional network of multivariate Hawkes processes in the presence of hidden confounders. Our method extends the trim regression \citep{cevid2020spectral} to the setting of point process data. The choice of trim regression as the starting point was motivated by the fact that its assumptions are less stringent than conditions required for the alternative HIVE procedure, especially for a stable point process network with dense confounding effects. Empirically, our procedure shows superior edge-selection performance compared with HIVE and a na\"ive method that ignores the unobserved nodes.

Our estimates assume a linear Hawkes process with a particular parametric form of the transition function. Thus, the proposed method identifies causal effects only if these modeling assumptions are valid. When the modeling assumptions are violated, the estimated effects may not be causal. In other words, the method is primarily designed to generate causal hypotheses---or facilitate \textit{causal discovery}---and the results should be  interpreted with caution. 
Extending the proposed approach to model the transition function nonparametrically and learn its form adaptively from data would thus be an important future research direction. 
In addition, given that non-linear link functions are often used when analyzing spike train data \citep{PANINSKI2007, Pillow2008}, it would also be of interest to develop casual-estimation procedure for non-linear Hawkes processes.

\bibliography{ref}
\appendix
\clearpage
%%%%%%%%%%%%%
\section{Additional Details on HIVE}\label{sec:hive}
%%%%%%%%%%%%%
%
We introduce additional notations before illustrating the method.

Let $Y(t) = \left(  Y_1(t), \dots, Y_p(t) \right)^\top$, $X(t) = \left(  x_1(t), \dots, x_p(t) \right)^\top$, $Z(t) = \left(  z_1(t), \dots, z_q(t) \right)^\top$ and $E(t) = \left(  \epsilon_1(t), \dots, \epsilon_p(t) \right)^\top$. 
Then, we rewrite \eqref{eq:hidden_var_model} simultaneously for all components:
\begin{align}
\label{eq:hive_model} 
    Y(t) = \bm{\mu} + \Theta X(t)  + \Delta Z(t) +   E(t), 
\end{align}
where $\Theta = \mat{  \bm{\beta}^\top_1 \\ \dots \\  \bm{\beta}^\top_p } \in \mathbf{R}^{p \times p}$ and $\Delta = \mat{ \bm{\delta}^\top_1 \\ \dots \\ \bm{\delta}^\top_p } \in \mathbf{R}^{p \times q}$ are connectivity matrix between the observed and unobserved components, respectively. $\bm{\mu}= \left( \mu_1, \dots, \mu_p \right)^\top \in \mathbf{R}^{p}$ is the vector of spontaneous rate. 

To illustrate the confounding induced by the hidden process, we project $Z(t)$ onto the space spanned by $X(t)$ as 
\begin{align}
    Z(t) = \bm{\nu} +  A X(t)  + W(t), 
\end{align}
where $A$ is the projection matrix, representing the cross-sectional correlation between $Z$ and $X$. Then, \eqref{eq:hive_model}  becomes
\begin{align} \label{eq:hive_formula}
     Y(t) =\widetilde{\bm{\mu} } + \widetilde{\Theta}  X(t) 
    + \widetilde{E}(t),
\end{align}
where 
\begin{align*}
  \widetilde{\bm{\mu} } &=  \bm{\mu} + \Delta \bm{\nu} ,\\
  \widetilde{\Theta} &= \Theta +  \Delta A, \\
  \widetilde{E}(t)   &=  E(t) + \Delta  W(t) . 
\end{align*}
From the above, it is easy to see that the correlations between the observed and unobserved processes determine the strength the confounding. Specifically, unless $A = 0$ ---i.e., when the observed and unobserved processes are independent, directly regressing $Y(t)$ on $X(t)$ produces biased estimates on $\Theta$. Under the condition that $\Theta \perp \Delta$---i.e., the column space of $\Theta$ is orthogonal to the column space of $\Delta$, HIVE gets around this issue by finding a projection matrix, $P_{\Delta^\perp}$, that projects $\Delta$ onto its orthogonal space ---i.e., $ P_{\Delta^\perp} \Delta = 0$. Moreover, because of the orthogonality assumption, $ P_{\Delta^\perp} \Theta = \Theta$. Therefore, when multiplying both sides in \eqref{eq:hive_model} by $ P_{\Delta^\perp} $, the unobserved term disappears. Specifically, letting $\widetilde{Y}(t) = P_{\Delta^\perp} Y(t) $, \eqref{eq:hive_model} becomes 
\begin{align}
\label{eq:hive_model_proj} 
    \widetilde{Y}(t) =  P_{\Delta^\perp}  \bm{\mu} + \Theta  X(t)  +    P_{\Delta^\perp}  E(t) .
\end{align}
Consequently, regressing $\widetilde{Y}(t)$ on $X(t)$ produces unbiased estimates on $\Theta$ (using penalized regression with $\ell_1$-penalty on $\Theta$ under the high-dimensional setting when $p$ is allowed to grow with the sample size $T$). In order to obtain $P_{\Delta^\perp}$, HIVE first calculates $\widetilde{E}(t) $ in \eqref{eq:hive_formula} and then implement \textit{heteroPCA} algorithm \citep{zhang2019heteroskedastic} to estimate the latent column space of $\Delta$ thus to obtain $P_{\Delta}$. Then, the method obtains the corresponding orthogonal project as $P_{\Delta^\perp} = I -  P_{\Delta} $. We refer the interested readers to \citet{bing2020adaptive} for details about the method. 

\clearpage
%%%%%%%%%%%%%%%%%%%%%%%%%%%%%%%%%%%%%%%%%%%%%%%%%%%%%
\section{Proof of Main Results}\label{sec:proofs}
Since our focus is on the estimation error for $\bm{\beta}_i$, we consider the perturbation model in \eqref{eq:perturb} in the following.

Let $\bm{\theta}_i = \mat{\mu_i & \bm{\beta}_i }^\top$ be the true model parameter and $\widehat{\bm{\theta}}_i = \mat{ \widehat{\mu}_i & \widehat{\bm{\beta}}_i }^\top$ be the optimizer for \eqref{eq:lasso}.
Recall that the set of active indices, $S_i= \{j:  \beta_{ij}\ne 0 , 1\le j \le p \}$, and $s_i = |S_i|$ and $s^* \equiv \max_{ 1\le i \le p} s_i $. 
Because optimization problem \eqref{eq:lasso} can be solved separately for each component process, in the follows we focus on the estimation consistency for one component process. For ease of notation, we drop the subscript $i$; that is, we use $\bm{x}(t)$ for $\bm{x}_i(t)$, $\bm{\theta}$ for $\bm{\theta}_i$, $dN(t)$ for $dN_i(t)$,  $\lambda(t)$ for $\lambda_i(t)$, $\bm{b}$ for $\bm{b}_i$, $S$ for $S_i$ and $\widetilde{S}$ for $\widetilde{S}_i$. 

Next, we state two lemmas that will be used in the proof of main results. 

\begin{Lemma}[\citet{vandegeer1995}]
	\label{lemma_vandergeer1995}
	Suppose there exists $\lambda_{\max} $ such that $\lambda(t) \le \lambda_{\max}$ where  $\lambda(t)$ is the intensity function of Hawkes process defined in 
	\eqref{eq:linear_hawkes}. Let $H(t)$ be a bounded function that is $\mathcal{H}_t$-predictable.
	Then, for any $\epsilon > 0$,
	\begin{align*}
	\frac{1}{T} \int_0^T H(t) \bigg \{ \lambda(t) dt - dN(t) \bigg \}
	\le 4 \bigg \{ \frac{\lambda_{\max} }{2T }   \int_0^T H^2(t) dt \bigg \}  ^{1/2} \epsilon^{1/2},
	\end{align*}
	with probability at least $1 - C\exp(-\epsilon T)$, for some constant $C$.
\end{Lemma}

\begin{Lemma}[\citet{wang2020statistical}]
\label{lemma_min_eigen}
Suppose the Hawkes process defined in \eqref{eq:linear_hawkes}
satisfies Assumptions~\ref{assumption1}--~\ref{assumption4}. 
Let $\textrm{Q} =  \frac{1}{T}\int_0^{T}   \mat{1 \\ \bm{x}(t) } 
\mat{1 &  \bm{x}^\top(t)} dt$, where $\bm{x}(t)$ is defined in \eqref{eq:design_column_xt}. Then, there exists $\gamma_{\max}  
\ge \gamma_{\min} >0$ such that
\begin{align*}
\gamma_{\max} \ge \Lambda_{\max} \left( \textrm{Q}  \right)  \ge  \Lambda_{\min}\left( \textrm{Q}  \right) \ge \gamma_{\min} > 0,
\end{align*}
with probability at least $1-c_1 p^2 T \exp(-c_2 T^{1/5})$, where constants
$c_1, c_2$ depending on the model parameters and the transition kernel.
\end{Lemma}

\hfill %\break

%%%%%%%%%%%%%%%%%%%%%%%%%%%%%%%%%
% proof of theorem 1
%%%%%%%%%%%%%%%%%%%%%%%%%%%%%%%%%%
\noindent
\textbf{Proof of Theorem~\ref{theorem1} }:
While the skeleton of the proof follows from \citet[][Theorem~2]{cevid2020spectral}, the following two conditions are needed because of the Hawkes process data's unique dependency structure.

\begin{condition}\label{def:RSC}
	There exist constants $\gamma_{\min}, c, C >0 $ such that
	\begin{align*}
	\mathbb{P}\left(
	\min_{\Delta \in 
	\mathcal{C}(L, S) } 
	\frac{1}{T}    \left \lVert  \widetilde{X} \Delta \right \rVert_2^2  
	  \ge  \gamma_{\min} \lVert \Delta \rVert^2_2 
	\right) \ge 1-c p^2 T \exp(- C T^{1/5}) ,
	\end{align*}
		where 
	$  \mathcal{C}(L, S)=
	\{ \bm{\alpha} : \lVert \bm{\alpha}_{S^c} \rVert_1 \le L \lVert \bm{\alpha}_S \rVert_1 \}$.
\end{condition}
Condition~\ref{def:RSC} is referred as the restrict strong convexity (RSC) \citep{Negahban2012}. Lemma~\ref{lemma_min_eigen} by \citet{wang2020statistical} has shown Condition~\ref{def:RSC} holds when $\widetilde{X} = X$ under Assumption~\ref{assumption1}-~\ref{assumption4}. Since the min eigenvalue of $\widetilde{X}$ stays the same with our choice of $F$, Condition~\ref{def:RSC} holds for $\widetilde{X} = FX$. 

\begin{condition}\label{def:tail_bound}
There exist $c,C >0$ such that 
\begin{align*}
\mathbb{P}\left( 
\frac{1}{T}
\left \lVert  \widetilde{X} \nu \right \rVert_\infty 
\le  
C \Lambda^2_{\max}\left(F\right) T^{-2/5}  
 \right) \ge 1- c p \exp(-T^{1/5} ) ,
\end{align*}
where $\nu$ is defined in \eqref{eq:perturb}.
\end{condition}
Condition~\ref{def:tail_bound} holds as a result of Lemma~\ref{lemma_vandergeer1995} by \citet{vandegeer1995}.

Under the two conditions, we achieve the conclusion as follows.
Because $\widehat{\bm{\theta}}$ is the optimizer for \eqref{eq:lasso}, 
\begin{align*}
    \frac{1}{T} \lVert  \widetilde{Y} - \widetilde{X}\widehat{\bm{\theta}} \rVert^2_2 
    + \lambda \lVert \widehat{\bm{\beta}} \rVert_1 
    &\le 
     \frac{1}{T} \lVert  \widetilde{Y} - \widetilde{X}\bm{\theta} \rVert^2_2 
    + \lambda \lVert \bm{\beta} \rVert_1 \\
    \frac{1}{T} \left \lVert  \widetilde{X}\left( \widehat{\bm{\theta}}  - \bm{\theta} - \bm{b} \right)  \right \rVert^2_2 
    + \lambda \lVert \widehat{\bm{\beta}} \rVert_1 
    &\le 
     \frac{2}{T} \int_{t=0}^T \nu(t) \widetilde{X}(t) \left(  \widehat{\bm{\theta}}  - \bm{\theta}  \right) 
     + \frac{1}{T} \lVert \widetilde{X}\bm{b}  \rVert^2_2
    + \lambda \lVert \bm{\beta} \rVert_1
\end{align*}

Under Condition~\ref{def:tail_bound}, 
\begin{align*}
\frac{2}{T} \int_{t=0}^T \nu(t) \widetilde{X}(t) \left(  \widehat{\bm{\theta}}  - \bm{\theta}  \right) 
  \le 
\frac{2}{T} \left \lVert \int_{t=0}^T \nu(t) \widetilde{X}(t) \right \rVert_\infty
\left 
\lVert   \widehat{\bm{\theta}}  - \bm{\theta} 
\right 
\rVert_1 
\le   \psi    \left 
\lVert   \widehat{\bm{\theta}}  - \bm{\theta}
\right 
\rVert_1  ,
\end{align*}
with probability at least $1- c_1 p \exp(-T^{1/5} )$,
where $ \psi = C_1 \Lambda^2_{\max}\left(F\right) T^{-2/5} $.

Letting $\bm{\theta}_S = \mat{ u & \bm{\beta}_S}^\top$ and 
$\bm{\theta}_{S^c} = \mat{ u & \bm{\beta}_{S^c}}^\top$,
\begin{align*}
    \frac{1}{T} \left \lVert  \widetilde{X}\left( \widehat{\bm{\theta}}  - \bm{\theta} - \bm{b} \right)  \right \rVert^2_2 
    + \lambda \lVert \widehat{\bm{\beta}} \rVert_1 
    &\le 
      \psi    \left 
\lVert   \widehat{\bm{\theta}}  - \bm{\theta}
\right 
\rVert_1 
     + \frac{1}{T} \lVert \widetilde{X}\bm{b}  \rVert^2_2
    + \lambda \lVert \bm{\beta} \rVert_1 \\
    \frac{1}{T} \left \lVert  \widetilde{X}\left( \widehat{\bm{\theta}}  - \bm{\theta} - \bm{b} \right)  \right \rVert^2_2 
    + (\lambda - \psi) 
    \lVert \widehat{\bm{\theta}}_{S^c} - \bm{\theta}_{S^c}\rVert_1
    &\le 
      (\lambda + \psi) 
    \left \lVert \widehat{\bm{\theta}}_S - \bm{\theta}_S  \right \rVert_1
    + \frac{1}{T} \lVert \widetilde{X}\bm{b}  \rVert^2_2
\end{align*}

Next, we discuss in two conditions: i) $\frac{1}{T} \lVert \widetilde{X}\bm{b}  \rVert^2_2 \le \lambda 
    \left \lVert \widehat{\bm{\theta}}_S - \bm{\theta}_S  \right \rVert_1$ and ii) 
    $\frac{1}{T} \lVert \widetilde{X}\bm{b}  \rVert^2_2 \ge \lambda 
    \left \lVert \widehat{\bm{\theta}}_S - \bm{\theta}_S  \right \rVert_1$.
    
First, when  $\frac{1}{T} \lVert \widetilde{X}\bm{b}  \rVert^2_2 \le \lambda 
    \left \lVert \widehat{\bm{\theta}}_S - \bm{\theta}_S  \right \rVert_1$, 
\begin{align*}
   \frac{1}{T} \left \lVert  \widetilde{X}\left( \widehat{\bm{\theta}}  - \bm{\theta} - \bm{b} \right)  \right \rVert^2_2 
    + (\lambda - \psi) 
   \lVert \widehat{\bm{\theta}}_{S^c} - \bm{\theta}_{S^c}\rVert_1
    &\le 
      (2\lambda + \psi) 
    \left \lVert \widehat{\bm{\theta}}_S - \bm{\theta}_S  \right \rVert_1.
\end{align*}

The above implies 
\begin{align*}
    (\lambda - \psi) 
   \lVert \widehat{\bm{\theta}}_{S^c} - \bm{\theta}_{S^c}\rVert_1
   \le (2\lambda + \psi) 
    \left \lVert \widehat{\bm{\theta}}_S - \bm{\theta}_S  \right \rVert_1 , 
\end{align*}
which means $  \widehat{\bm{\alpha}}_{S^c} - \bm{\alpha}_{S^c} \in \mathcal{C}(L, S)=\{ \bm{\alpha} : \lVert \bm{\alpha}_{S^c}\rVert_1 \le L \lVert \bm{\alpha}_S \rVert_1 \}$ for $L = \frac{2\lambda + \psi}{\lambda  - \psi}$. 

Taking $\lambda = 2\psi$,   
\begin{align*}
   & \frac{1}{T} \left \lVert  \widetilde{X}\left( \widehat{\bm{\theta}}  - \bm{\theta} - \bm{b} \right)  \right \rVert^2_2 
    + (\lambda - \psi) 
   \left \lVert \widehat{\bm{\theta}} - \bm{\theta} \right \rVert_1 \\
  \le   &
      3\lambda
      \sqrt{s^*} 
    \lVert \widehat{\bm{\theta}}_S - \bm{\theta}_S \rVert_2 \\
    \le & 3\lambda
      \sqrt{s^*} 
    \frac{1}{ \gamma_{\min} \sqrt{T}} 
    \left \lVert \widetilde{X}\left(  \widehat{\bm{\theta}} - \bm{\theta} \right) \right \rVert_2 \\
\le &
3\lambda
      \sqrt{s^*} 
    \frac{1}{ \gamma_{\min} \sqrt{T}}
\left\{ 
    \left \lVert \widetilde{X}\left(  \widehat{\bm{\theta}} - \bm{\theta} 
    - \bm{b} \right) \right \rVert_2 +
    \left \lVert  \widetilde{X}\bm{b} \right \rVert_2   
\right\}
 \\
\le &
3\lambda
      \sqrt{s^*} 
    \frac{1}{ \gamma_{\min} \sqrt{T}}
    \left \lVert \widetilde{X}\left(  \widehat{\bm{\theta}} - \bm{\theta} 
    - \bm{b} \right) \right \rVert_2 +
 3\lambda
      \sqrt{s^*} 
    \frac{1}{ \gamma_{\min} \sqrt{T}}   \left \lVert  \widetilde{X}\bm{b} \right \rVert_2   \\
\le &
\frac{9}{2}\lambda^2
      s^* \frac{1}{ \gamma_{\min}^2 }
+ \frac{1}{ T }
    \left \lVert \widetilde{X}\left(  \widehat{\bm{\theta}} - \bm{\theta} 
    - \bm{b} \right) \right \rVert^2_2
+     
  \frac{1}{ T}   \left \lVert  \widetilde{X}\bm{b} \right \rVert^2_2,
\end{align*}
where the second inequality is by Condition~\ref{def:RSC} and the last step is by using $xy \le \frac{1}{4}x^2 + y^2$ twice. Therefore, we get 
\begin{align*}
(\lambda - \psi) \left \lVert \widehat{\bm{\theta}} - \bm{\theta} \right \rVert_1
  \le &
\frac{9}{2}\lambda^2
      s^* \frac{1}{ \gamma_{\min}^2 }
+     
  \frac{1}{ T}   \left \lVert  \widetilde{X}\bm{b} \right \rVert^2_2.  
\end{align*}

When $\frac{1}{T} \lVert \widetilde{X}\bm{b}  \rVert^2_2 \ge \lambda 
    \left \lVert \widehat{\bm{\theta}}_S - \bm{\theta}_S  \right \rVert_1$,
    \begin{align*}
   \frac{1}{T} \left \lVert  \widetilde{X}\left( \widehat{\bm{\theta}}  - \bm{\theta} - \bm{b} \right)  \right \rVert^2_2 
    + (\lambda - \psi) 
   \left \lVert \widehat{\bm{\theta}} - \bm{\theta} \right \rVert_1 
   \le  \frac{3}{T} \left \lVert  \widetilde{X}\bm{b} \right \rVert^2_2. 
    \end{align*}
    
Combining the two cases, we always have
\begin{align*}
(\lambda - \psi) \left \lVert \widehat{\bm{\theta}} - \bm{\theta} \right \rVert_1
  \le &
\frac{9}{2}\lambda^2
      s^*\frac{1}{ \gamma_{\min}^2 }
+     
  \frac{3}{ T}   \left \lVert  \widetilde{X}\bm{b} \right \rVert^2_2.  
\end{align*}

Thus, taking $\lambda = 2\psi = O(\Lambda^2_{\max}\left( F\right) T^{-2/5})$ and dividing both sides by $\frac{1}{2}\lambda$, we achieve the conclusion that 
\begin{align*}
 \left \lVert \widehat{\bm{\theta}} - \bm{\theta} \right \rVert_1
  \le &
C_1  \Lambda^2_{\max}(F) \frac{s^*}{ \gamma_{\min}^2 } T^{-2/5} 
+     
  C_2 T^{-3/5} \Lambda^{-2}_{\max}(F) \left \lVert  \widetilde{X}\bm{b} \right \rVert^2_2.  
\end{align*}

\hfill

%\newpage
\noindent
\textbf{Proof of Corollary~\ref{corollary1} }:
Notice that 
\begin{align*}
     \frac{1}{T}\lVert \widetilde{X} b \rVert^2_2 \le   \Lambda^2_{\max}\left( F \right)  \frac{1}{T} \lVert X b \rVert^2_2 \le  \Lambda^2_{\max}\left( F \right)  \gamma_{\max} \lVert b \rVert^2_2 , 
\end{align*}
with probability at least $1-c_1 p^2 T \exp(-c_2 T^{1/5})$, 
where the second inequality is by Lemma~\ref{lemma_min_eigen}. 

Then, Corollary~\ref{corollary1} is a direct result from Theorem~\ref{theorem1} by plugging in $\lVert b \rVert^2_2$.

%\clearpage 
\hfill %\break

%%%%%%%%%%%%%%%%%%%%%%%%%%%%%%%%%
% proof of theorem 3
%%%%%%%%%%%%%%%%%%%%%%%%%%%%%%%%%%
\noindent
\textbf{Proof of Theorem~\ref{theorem2}}:
Recall $S = \{ \beta_{ij}: \beta_{ij} \ne 0, 1\le i,j\le p \}$ and $S_C = \{ \beta_{ij}: \beta_{ij} = 0, 1\le i,j\le p \}$ .
To establish selection consistency, we need two parts. 
First, we show that our estimates on the true zero and non-zero coefficients can be separated with high probability; 
that is, there exists some constant $\Delta>0$ such that for $\beta_{S} \in S $ and $\beta_{S_C} \in S_C$,
$| \widehat{\beta}_{S} - \widehat{\beta}_{S_C}| \ge \Delta $ with high probability. 
By the $\beta$-min condition specified in Assumption~\ref{assumption5},
we have  $\beta_{ij} \in S \ge 2\tau$. Theorem~\ref{theorem1} shows that
 for $1\le i,j \le p$, 
$ | \widehat{\beta}_{ij} - \beta_{ij} | \le \tau $ 
with probability at least $1-c_1 p^2  T \exp(-c_2 T^{1/5})$. Then, 
for any $\beta_{S} \in S $ and $\beta_{S_C} \in S_C$,
\begin{align*}
| \widehat{\beta}_{S} - \widehat{\beta}_{S_C}|  
&= | \widehat{\beta}_{S} -\beta_{S} - ( \widehat{\beta}_{S_C} -\beta_{S_C}  ) + \beta_{S} - \beta_{S_C} |  \\
&\ge 
| \beta_{S}  - \beta_{S_C} | -  | \widehat{\beta}_{S} -\beta_{S} | - | \widehat{\beta}_{S_C} -\beta_{S_C} |  \\
&\ge \beta_{min} - 2\tau .
\end{align*}
This means the estimates on zero and non-zero coefficients can be separated with high probability.

Next, we show there exists a post-selection threshold that allows to correctly identify $S$ and $S_C$ based on the estimates. In fact, the post-selection estimator is 
$$ 
\widetilde{\beta}  = \widehat{\beta} \mathbf{1}( |\widehat{\beta}| >  \tau) .
$$
By Theorem~\ref{theorem1}, 
we have $|\widehat{\beta}_{S_C}| \le \tau $, with probability $1-c_1 p^2  T \exp(-c_2 T^{1/5})$. 
Then,
$$ 
\widetilde{\beta}_{S_C} = \widehat{\beta}_{S_C} \mathbf{1}(\widehat{\beta}_{S_C} >   \tau_S) = 0,
$$ 
which means $\widetilde{\beta}$ selects $\beta_{S_C} $ into $S_C$ with high probability.
In addition, since $| \widehat{\beta}_{S}  -  \beta_{S} | \le \tau $,
$$
|  \widehat{\beta}_{S}  | \ge | \beta_{S}| - \tau \ge \beta_{min} - \tau > \tau  > 0.
$$
Therefore, 
$$ 
\widetilde{\beta}_{S} = \widehat{\beta}_{S}  \mathbf{1}(|\widehat{\beta}_{S}|  >   \tau) = \widehat{\beta}_{S} \ne 0 ,
$$
which means $\widetilde{\beta}_{S }$ selects $\beta_{S}$ into $S$ with high probability. 
 
Combining the two sides, the post-selection estimator $\widetilde{\beta}$ identifies $S$ and $S_C$ with high probability.

\end{document}